\documentclass[12pt,draftclsnofoot,onecolumn,letterpaper]{IEEEtran}
\IEEEoverridecommandlockouts
\usepackage{mathrsfs}
\usepackage{amsfonts}
\usepackage{ifpdf}
\usepackage{cite}
\ifCLASSINFOpdf
\usepackage[pdftex]{graphicx}
\else \fi
\usepackage{amsmath}
\usepackage{array}
\usepackage{mdwmath}
\usepackage{mdwtab}
\usepackage{eqparbox}
\usepackage[tight,footnotesize]{subfigure}
\usepackage{algorithmic}
\usepackage{algorithm}
\usepackage{amsmath}
\usepackage{epsfig}
\usepackage{ae}
\usepackage{amssymb}
\usepackage{times}
\usepackage{amsmath}   % add1
\usepackage{booktabs}  % add2
\usepackage{threeparttable} %add3
\usepackage{multirow}       %add4
\usepackage{xcolor}

\usepackage{amsmath,amssymb,amsfonts}
\usepackage{graphicx}
\usepackage{textcomp}
\usepackage{xcolor}

\def\BibTeX{{\rm B\kern-.05em{\sc i\kern-.025em b}\kern-.08em
    T\kern-.1667em\lower.7ex\hbox{E}\kern-.125emX}}

\begin{document}
%----------------------------------------Make Title -----------------------------------------

%\title{On technical contributions of R-learning based deep reinforcement learning algorithm design}

\title{Optimizing the Long-Term Average Reward for Continuing MDPs: A Technical Report}

%\title{Service Selection and Resource Allocation in Multi-tier Heterogeneous Cellular Networks: A load balancing perspective}

%-------------------------------------------------------------------------------------------
%\author{\IEEEauthorblockN{Chao Xu$^{\dag,\S}$, Xijun Wang$^*$, Howard H. Yang$^{\ddag}$, Hongguang Sun$^{\dag,\S}$, and Tony Q. S. Quek$^{\ddag}$}
%\IEEEauthorblockA{$^{\dag}$School of Information Engineering, Northwest A\&F University, Yangling, Shaanxi, China\\
%$^{\S}$Key Laboratory for Agricultural Internet of Things, Ministry of Agriculture and Rural Affair, Yangling, China\\
%$^*$School of Electronics and Communication Engineering, Sun Yat-sen University, Guangzhou, China\\
%$^{\ddag}$Information System Technology and Design Pillar, Singapore University of Technology and Design, Singapore}
%
%\thanks{
%This paper is supported by National Natural Science Foundation of China (61701372), Talents Special Foundation of Northwest A \& F University (Z111021801), Key Research and Development Program of Shaanxi (2019ZDLNY07-02-01), Fundamental Research Funds for the Central Universities of China(SYSU: 19lgpy79), and Research Fund of the Key Laboratory of Wireless Sensor Network \& Communication (Shanghai Institute of Microsystem and Information Technology, Chinese Academy of Sciences) under grant 20190912.
%}
%
%}

%
\author{
%Chao Xu, Yiping Xie, Xijun Wang, \\ Howard H. Yang, Dusit Niyato, and Tony Q. S. Quek

Chao Xu, {\em Member, IEEE}, Yiping Xie, \\ Xijun Wang, {\em  Member, IEEE}, Howard H. Yang, {\em Member, IEEE}, \\Dusit Niyato, {\em Fellow, IEEE}, and Tony Q. S. Quek, {\em Fellow, IEEE}

\vspace{-0.2cm}
}

\maketitle
%---------------------------------------Make Abstract--------------------------------------
\begin{abstract}
Recently, we have struck the balance between the information freshness, in terms of age of information (AoI), experienced by users and energy consumed by sensors, by appropriately activating sensors to update their current status in caching enabled Internet of Things (IoT) networks \cite{Our_TWC_2021}. To solve this problem, we cast the corresponding status update procedure as a continuing Markov Decision Process (MDP) (i.e., without termination states), where the number of state-action pairs increases exponentially with respect to the number of considered sensors and users. Moreover, to circumvent the curse of dimensionality, we have established a methodology for designing deep reinforcement learning (DRL) algorithms to maximize (resp. minimize) the average reward (resp. cost), by integrating R-learning, a tabular reinforcement learning (RL) algorithm tailored for maximizing the long-term average reward, and traditional DRL algorithms, initially developed to optimize the discounted long-term cumulative reward rather than the average one. In this technical report, we would present detailed discussions on the technical contributions of this methodology.
\end{abstract}
%--------------------------------------------- Make Key words-----------------------------
\begin{IEEEkeywords}
Continuing MDP, deep reinforcement learning, long-term average reward, discounted long-term cumulative reward.
\end{IEEEkeywords}
%%%%%%%%%%%%%%%%%%%%%%%%%%%%%%%%%%%%%%%%%%%%%%%%%%%%%%%%%%%%%%%%%%%%%%%%%%%%%%%%%%%%%%%%%%%%%%%%%%%%%%%%%%%%%%%%%%%%%%%%%%%%%%%%
%%%%%%%%%%%%%%%%%%%%%%%%%%%%%%%%%%%%%%%%%%%%%%%%%%%%%%%%%%%%%%%%%%%%%%%%%%%%%%%%%%%%%%%%%%%%%%%%%%%%%%%%%%%%%%%%%%%%%%%%%%%%%%%%
%%%%%%%%%%%%%%%%%%%%%%%%%%%%%%%%%%%%%%%%%%%%%%%%%%%%%%%%%%%%%%%%%%%%%%%%%%%%%%%%%%%%%%%%%%%%%%%%%%%%%%%%%%%%%%%%%%%%%%%%%%%%%%%%

\section{Introduction}

Acting as a critical and integrated infrastructure, the Internet of Things (IoT) enables ubiquitous connection for billions of things in our physical world, ranging from tiny, resource-constrained sensors to more powerful smart phones and networked vehicles\cite{Survey_IoT_Applications_2015}. In general, the sensors are powered by batteries with limited capacities rather than fixed power supplies. Thus, to exploit the benefits promised by IoT networks, it is essential to well address the energy consumption issue faced by sensors. Recently, caching has been proposed as a promising solution to lower the energy consumption of sensors by reducing the frequency of environmental sensing and data transmission \cite{IoT_Caching_2016_Network,Caching_IoT_EH_ICC_2016,IoT_Caching_2017_RL}.

However, compared with the multimedia contents (e.g., music, video, etc.) in traditional wireless networks, the data packets in IoT networks have two distinct features: 1) The sizes of data packets generated by IoT applications are generally much smaller than those of multimedia contents. Therefore, for IoT networks, the storage capacity of each ECN is generally sufficient to store the latest status updates generated by all the sensors. 2) For many real-time IoT applications, the staleness of information at the user side can radically deteriorate the accuracy and reliability of derived decisions. As such, the main concern for edge caching enabled IoT networks would be related to how to properly update the cached data to lower the energy consumption of sensors and meanwhile improve the information freshness at users or, in other words, to refresh the cached items in a cost-efficient and timely fashion.

To this end, in our recent study \cite{Our_TWC_2021}, we focus on striking the balance between the information freshness, in terms of age of information (AoI), experienced by users and energy consumed by sensors, by appropriately activating sensors to update their current status in caching enabled IoT networks. To solve this problem, we cast the corresponding status update procedure as a continuing Markov Decision Process (MDP) (i.e., without termination states), where the number of state-action pairs increases exponentially with respect to the number of considered sensors and users. Moreover, to circumvent the curse of dimensionality, we have established a methodology for designing deep reinforcement learning (DRL) algorithms to maximize the average reward, by integrating R-learning \cite{RL_1993}, a tabular reinforcement learning (RL) algorithm tailored for  maximizing the long-term average reward, and available DRL algorithms, initially developed to optimize the discounted long-term cumulative reward rather than the average one. Using the established methodology, more R-learning based DRL algorithms can be devised for addressing various continuing decision-making tasks that aim at maximizing the long-term average reward. Interested readers are referred to our paper \cite{Our_TWC_2021} for the detailed problem formulation, algorithm design, as well as performance comparison between our proposed algorithm and baseline DRL algorithms.

In this report, we would like to discuss the technical contributions of the methodology established in \cite{Our_TWC_2021}, by comparing our proposed R-learning DRL algorithm with existing tabular RL and DRL algorithms. Particularly, we will first thoroughly analyze why, \textbf{in theory, the existing DRL algorithms, initially developed to optimize the discounted long-term cumulative reward, is not suitable for maximizing the long-term average reward}. Then, we will discuss why we choose to integrate \textbf{the R-learning \cite{RL_1993}, rather than the relative value iteration (RVI) Q-learning (another popular tabular RL algorithm) \cite{DP_OC_Introduction}, with traditional DRL algorithms when designing DRL algorithms to maximize the long-term average reward for continuing MDPs}, and more importantly summarize what modifications we have performed to make the proposed algorithm more compatible with the function approximation realized by utilizing artificial neural networks (ANNs).

\section{Maximizing the discounted long-term cumulative reward is not the same as maximizing the long-term average reward for continuing MDPs!}

For a continuing MDP with the state space $\mathbb S$, action space $\mathbb A$, and achievable reward set $\mathbb U$, the interaction between the agent and environment continues infinitely, i.e., there is no terminal state. Then, under the assumption that the controlled Markov chain for policy $\pi$ is unichain, the achieved long-term average reward exists and is independent of the initial state $s_1$, which can be expressed as follows \cite{RL_Introduction}:
\begin{align} \label{Eq:Ave_Rew_Def}
\hat U(\pi) & = \lim _{h \rightarrow \infty} \frac{1}{h} \sum_{t=1}^{h} \mathbb{E}\left[U_{t} \mid s_{1}, a_{1: t} \sim \pi\right] \\ \nonumber
&=\lim _{t \rightarrow \infty} \mathbb{E}\left[U_{t} \mid s_{1}, a_{1: t} \sim \pi\right] \\ \nonumber
&=\sum_{s \in \mathbb S} \mu_{\pi}(s) \sum_{a \in \mathbb A} \pi(a \mid s) \sum_{s^{\prime} \in \mathbb S, u \in \mathbb U} p\left(s^{\prime}, u \mid s, a\right) u
\end{align}
where $U_{t}$ denotes the instantaneous reward obtained in decision epoch $t$, $\pi(a \mid s)$ is the probability that the agent chooses action $a$ at state $s$ under policy $\pi $, and $p\left(s^{\prime}, u \mid s, a\right)$ denotes the probability that the reward-state
pair $(u, s^{\prime})$ is observed by the agent, if it performs action $a$ at state $s$. Additionally, $\mu_{\pi}(s)$ is the steady-state distribution of the Markov chain following policy $\pi$, i.e.,
\begin{align}
\mu_{\pi}(s) = \lim _{t \rightarrow \infty} \operatorname{Pr}\left\{S_{t} = s \mid a_{1: t-1} \sim \pi\right\}
\end{align}
under which, if the agent takes actions according to $\pi$, it remains in the same distribution, i.e.,
\begin{align} \label{Eq:Steady_State_Rela}
\mu_{\pi}(s) =  \sum_{s^{\prime} \in \mathbb S} \mu_{\pi}(s^{\prime}) \sum_{a \in \mathbb A}\pi(a \mid s^{\prime}) p (s \mid s^{\prime}, a)
\end{align}
where $p (s \mid s^{\prime}, a)$ denotes the probability of the transition from state $s^{\prime}$ to state $s$ by taking action $a$.

On the other hand, the discounted long-term cumulative reward (i.e., the traditional discounted state-value function) for each state $s \in \mathbb S$ can be expressed as \cite{RL_Introduction}
\begin{align} \label{Eq:Ave_Rew_State_Val}
V^{\gamma}_{\pi}(s) & = \mathbb{E}_{\pi}\left[\sum_{k=0}^{\infty} \gamma^{k} U_{t+k} \mid S_{t}=s\right] \\ \nonumber
&=\sum_{a \in \mathbb A} \pi(a \mid s) \sum_{s^{\prime} \in \mathbb S} \sum_{u \in \mathbb U} p\left(s^{\prime}, u \mid s, a\right)\left[u+\gamma \mathbb{E}_{\pi}\left[\sum_{k=0}^{\infty} \gamma^{k} U_{t+1+k} \mid S_{t+1}=s^{\prime}\right]\right] \\ \nonumber
&=\sum_{a \in \mathbb A} \pi(a \mid s) \sum_{s^{\prime}\in \mathbb S} \sum_{u \in \mathbb U} p\left(s^{\prime}, u \mid s, a\right)\left[u+\gamma V_{\pi}^{\gamma}\left(s^{\prime}\right)\right]
\end{align}
where $\gamma \in [0,1)$ denotes the discount factor. Accordingly, the relationship between the average discounted state-value $J(\pi)$ and the average reward $\hat U(\pi)$ in the steady states can be expressed as
\begin{align} \label{Eq:Ave_Rew_State_Val_Rela}
J(\pi) &=\sum_{s \in \mathbb S} \mu_{\pi}(s) V_{\pi}^{\gamma}(s) \\ \nonumber
&= \sum_{s\in \mathbb S} \mu_{\pi}(s) \sum_{a \in \mathbb A} \pi(a \mid s) \sum_{s^{\prime} \in \mathbb S} \sum_{u \in \mathbb U} p\left(s^{\prime}, u \mid s, a\right)\left[u+\gamma V_{\pi}^{\gamma}\left(s^{\prime}\right)\right] \\ \nonumber
& \mathop {\rm{ = }}\limits^{\left( a \right)} \hat U(\pi)+\sum_{s\in \mathbb S} \mu_{\pi}(s) \sum_{a \in \mathbb A} \pi(a \mid s) \sum_{s^{\prime}\in \mathbb S} \sum_{u \in \mathbb U} p\left(s^{\prime}, u \mid s, a\right) \gamma V_{\pi}^{\gamma}\left(s^{\prime}\right) \\ \nonumber
& \mathop {\rm{ = }}\limits^{\left( b \right)} \hat U(\pi)+\gamma \sum_{s^{\prime} \in \mathbb S} V_{\pi}^{\gamma}\left(s^{\prime}\right) \sum_{s\in \mathbb S} \mu_{\pi}(s) \sum_{a \in \mathbb A} \pi(a \mid s) p\left(s^{\prime} \mid s, a\right) \\ \nonumber
& \mathop {\rm{ = }}\limits^{\left( c \right)} \hat U(\pi)+\gamma \sum_{s^{\prime} \in \mathbb S} V_{\pi}^{\gamma}\left(s^{\prime}\right) \mu_{\pi}\left(s^{\prime}\right) \\ \nonumber
&=\hat U(\pi)+\gamma J(\pi) =\hat U(\pi)+\gamma \hat U(\pi)+\gamma^{2} J(\pi) \\ \nonumber
&=\hat U(\pi)+\gamma \hat U(\pi)+\gamma^{2} \hat U(\pi)+\gamma^{3} \hat U(\pi)+\cdots  \\ \nonumber
& =\frac{1}{1-\gamma} \hat U(\pi)
\end{align}
where (a) follows the definition of the long-term average reward given in (\ref{Eq:Ave_Rew_Def}), (b) holds due to the fact that
\begin{align}
\sum_{u \in \mathbb U} p\left(s^{\prime}, u \mid s, a\right) = p\left(s^{\prime} \mid s, a\right), \forall s^{\prime} \in \mathbb S
\end{align}
and (c) follows the feature of the steady-state distribution presented in (\ref{Eq:Steady_State_Rela}).

From (\ref{Eq:Ave_Rew_State_Val_Rela}), it can be seen that for one continuing MDP with a given discount factor $\gamma$, the policy $\pi^*$ that maximizes the average reward $\hat U(\pi)$ would be exactly the same as that of maximizing the average discounted state-value $J(\pi)$. Nevertheless, for discounted MDPs, the main objective of developing model-based or model-free RL algorithms is to find the policy satisfying the Bellman optimality equation, i.e., \cite{RL_Introduction}
\begin{align} \label{Eq:Opt_Bell_Fun}
V^{\gamma}_{\pi_{*}}(s)
&=\sum_{a \in \mathbb{A}} \pi_{*}(a \mid s) \sum_{s^{\prime} \in \mathbb{S}} \sum_{u \in \mathbb{U}} p\left(s^{\prime}, u \mid s, a\right)\left[u+\gamma V_{\pi_{*}}^{\gamma}\left(s^{\prime}\right)\right] \\ \nonumber
&=\max _{a \in \mathbb{A}} \sum_{s^{\prime} \in \mathbb{S}} \sum_{u \in \mathbb{U}} p\left(s^{\prime}, u \mid s, a\right)\left[u+\gamma V^{\gamma}_{\pi^*}\left(s^{\prime}\right)\right] \\ \nonumber
&=\max _{a \in \mathbb{A}} Q^{\gamma}_{\pi_{*}}(s, a), \forall s \in \mathbb{S}
\end{align}
where $Q^{\gamma}_{\pi}(s, a)$ is the action-value function associated with policy $\pi$.
By comparing (\ref{Eq:Ave_Rew_State_Val_Rela}) and (\ref{Eq:Opt_Bell_Fun}), \textbf{it can be seen that there is no guarantee that the policy maximizing $V^{\gamma}_{\pi}(s)$ for all states is identical to that of maximizing the average reward $\hat U(\pi)$}, since the steady-state distribution $\mu_{\pi}(s)$ incorporated in (\ref{Eq:Ave_Rew_State_Val_Rela}) is also determined by the adopted policy $\pi$. In fact, by resorting to the Abel and Ces\`{a}ro limits, it has been proved in the recent work \cite{Dis_Ave_Limt_2014} that, for continuing MDPs, only when the discount factor approaches 1, then for each policy $\pi$ the following equation holds, i.e.,
\begin{align}
\frac{1}{1-\gamma} \hat U(\pi) = \mathop {\lim }\limits_{\gamma \rightarrow 1} V_{\pi}^{\gamma}(s), \forall s \in \mathbb S.
\end{align}
\textbf{That is, in theory, the algorithms proposed for maximizing the discounted cumulative reward do not apply to maximizing the average reward, unless in the limiting regime where $\gamma \rightarrow 1$.}
However, in practice, available RL algorithms that learn to optimize the discounted value function become increasingly unstable as $\gamma$ increases to 1, since in this case the discounted cumulative reward goes to infinity. As such, it is extremely challenging to use such algorithms to derive the policy maximizing the long-term average reward \cite{2019_NeurIPS}.

Actually, while there are some available studies focusing on designing learning algorithms to optimize the average reward for continuing MDPs in the tabular case, i.e., the number of the state-action pairs is small, when the function approximation (e.g., realized by utilizing ANNs or even linear approximation) is implemented, a plenty of similar states (or state-action pairs) would be generalized and hence, the update of values for different states are coupled, which may result in that some essential principles for algorithm design in tabular MDPs (e.g., policy improvement theorem) cannot be realized \cite{RL_Introduction}. Therefore, how to design DRL algorithms to maximize the long-term average reward is still an open problem \cite{2019_NeurIPS}. In this light, although many existing researches aim at maximizing (resp. minimizing) the average reward (resp. cost) for various problems, they still use the DRL algorithm designed for optimizing the discounted long-term cumulative reward. Wherein, to achieve a better performance, the discount factor (treated as a hyperparameter) shall be sophisticatedly tuned, e.g., see the simulation setting in recent works \cite{DF_Tuning_1,DF_Tuning_2,DF_Tuning_3}, since, as discussed above, \textbf{these algorithms are initially developed to optimize the obtained discounted cumulative reward rather than the average one}.

\section{Why have we chosen R-learning for the DRL algorithm design and what modifications have we done?}

%This is mainly due to the fact that, with function approximation,and hence cannot be differentiated, which however is not a problem for MPDs with the tabular representation.

When we develop DRL algorithms for the dynamic status update problem in \cite{Our_TWC_2021}, we have initially considered two representative model-free RL algorithms, which were devised to optimize the average reward for continuing MDP in the tabular case, i.e., RVI Q-learning \cite{DP_OC_Introduction} and R-leaning \cite{RL_1993}. Particularly, for the RVI Q-leaning, in iteration $k+1$, if the agent at state $s$ selects an action $a$ and observes the reward-state pair $(u,s^{\prime})$, then the action value function associated with each state-action pair $(s,a)$, denoted by $Q(s,a)$, is updated as
\begin{align} \label{Eq:RVI_Q_Learning}
Q(s,a)(k+1) = Q(s,a)(k) +\alpha \left(u+ \max _{a^{\prime} \in \mathbb{A}} Q(s^{\prime}, a^{\prime})(k) - \max _{\bar a \in \mathbb{A}} Q(s_{ref}, \bar a)(k) -  Q(s,a)(k) \right)
\end{align}
where $\alpha$ is the adopted learning rate, and $s_{ref}$ denotes a special state, which can be chosen arbitrarily but remains fixed during the entire training process. While there are a few recent efforts (e.g., \cite{DRL_Updating_2020} and \cite{RVI_Q_Pappas}) trying to devise DRL algorithms based on the RVI Q-leaning to maximize (resp. minimize) the average reward (resp. cost) for solving engineering issues in wireless networks, we argue that this methodology is fundamentally not well compatible with the large-scale RL in continuing MDPs. This is mainly attributed to the fact that in this case, by introducing function approximation, the states are generalized and have correlated value functions, and more importantly, the majority of them even cannot be visited during the training. As such, \textbf{the chosen special state significantly affects the learning efficiency and final results.} To this end, the traditional DRL algorithms, originally designed for optimizing the discounted long-term cumulative reward, are still widely adopted in recent studies (e.g.,\cite{DF_Tuning_1,DF_Tuning_2,DF_Tuning_3}) to optimize the average reward or cost, where the discount factor shall be sophisticatedly tuned to achieve a better performance.

\begin{figure} [!t]
\centering
\leavevmode \epsfxsize=4.5in  \epsfbox{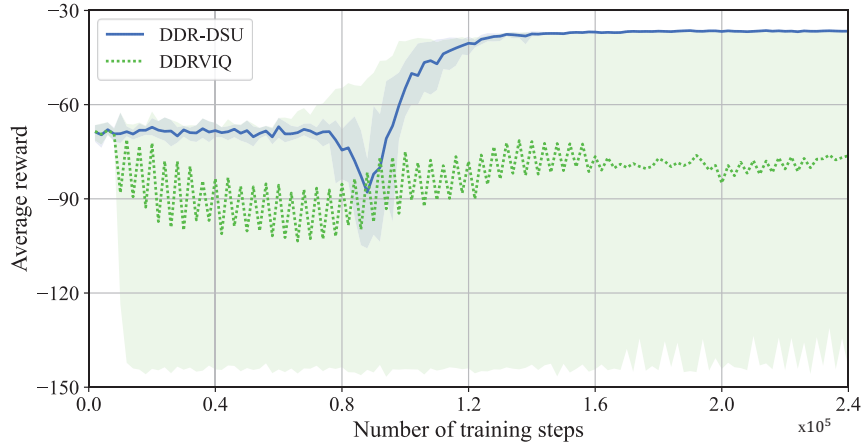}
\centering \caption{Convergence comparison of DDRVIQ and DDR-DSU, where the number of users is set as $N=24$ and $\beta_1 = \beta_2 =1$.} \label{Fig:Convergence_Ref}
\end{figure}

To demonstrate this, we have also developed a DRL algorithm by combining RVI Q-leaning with dueling deep Q-network, termed as DDRVIQ, and conducted the simulations with the same simulation parameters and ANN architecture as presented in \cite{Our_TWC_2021}. Here, identical to the setting of Fig. 4 in \cite{Our_TWC_2021}, we consider a network consisting of $K=8$ sensors and $N=24$ users, and set $\Delta_{\max}=10K(D_u+D_d)=160$.
During the simulation, for each random seed, we run 5 independent simulations, where the associated 5 special states $s_{ref}$ are randomly chosen from 5 disjoint subspaces to make the sampling more uniform. Particularly, for the $i$-th subspace ($\forall i \in \{1, 2, 3, 4, 5\}$), the AoI value belongs to $[(i-1)*\frac{\Delta_{max}}{5}+1, i*\frac{\Delta_{max}}{5}]$, i.e., there are $\left(\frac{\Delta_{\max}}{5}\right)^{K(N+1)}=32^{200}$ candidate special states in each subspace. The simulation results are shown in Fig. \ref{Fig:Convergence_Ref}, which are obtained by averaging over the independent runs for 6 random seeds (i.e., totally 30 runs). Wherein, the darker (solid or dotted) line shows the average value over runs and the shaded area
is obtained by filling the interval between the maximum and minimum values over runs. Besides, the performance of the DRL algorithm proposed in our work \cite{Our_TWC_2021} (termed as DDR-DSU) with the same seeds is also presented for comparison. It can be seen from Fig. \ref{Fig:Convergence_Ref} that the choice of $s_{ref}$ dramatically impacts the performance of DDRVIQ, which results in a less stable performance compared with our proposed DDR-DSU. As shown by the shaded area of DDRVIQ, the achieved average reward fluctuates significantly when different special states are chosen, while by implementing our proposed DDR-DSU, the much better convergence and stability can be obtained. Specifically, as presented in Table II of our study \cite{Our_TWC_2021}, for DDR-DSU, the mean and standard deviation of the achieved average reward during the last 10 evaluations are -36.59 and 0.19, respectively.

%In each independent run of simulation, the special state $S_{ref}$ is randomly chosen from in each subspace.
%
%Meanwhile, we divide the whole sate space into 5 subspace where in the $i$-th subspace the maximum AoI is as $i*\frac{\Delta_{max}}{5}$ and, in each independent run of simulation, the special state $S_{ref}$ is randomly chosen in each subspace. That is, for one random seed, we have conducted 5 simulations, where the special states are chosen from the 5 subspace, respectively.
%%
%%Accordingly, the size of the state space is $\left| {\mathbb S} \right| = \Delta_{max}^{K*(N+1)}=160^{200}$.
%%
%%We

For R-learning \cite{RL_1993}, in the $(k+1)$-th iteration, the value function associated with each state-action pair $(s,a)$, denoted by $R(s,a)(k+1)$, is updated according to
\begin{align} \label{Eq:R_Learning}
R(s,a)(k+1) = R(s,a)(k) +\alpha_R \left(u-\tilde{U}(k)+ \max _{a^{\prime} \in \mathbb{A}} R(s^{\prime}, a^{\prime})(k) -  R(s,a)(k) \right)
\end{align}
where $\tilde{U}(k)$ denotes the estimated average reward, and $\alpha_R $ the adopted learning rate. During each iteration, after (\ref{Eq:R_Learning}) is completed, $\tilde{U}(k)$ is updated as shown in (\ref{Eq:R_Learning_Ave_R})
\begin{align} \label{Eq:R_Learning_Ave_R}
&\tilde{U}(k+1) = \\ \nonumber
&\begin{cases}
 \tilde{U}(k) +\alpha_U \left(u +  \mathop {\max }\limits_{a^{\prime} \in \mathbb{A}} R(s^{\prime}, a^{\prime})(k+1) -  R(s,a)(k+1) - \tilde{U}(k) \right), & \text{{$R(s,a)(k+1) = \mathop {\max }\limits_{a^{\prime} \in \mathbb{A}} R(s, a^{\prime})(k+1)$}}\\
\tilde{U}(k), & \text{otherwise}
\end{cases}
\end{align}
where $\alpha_U$ is the adopted learning rate, and the condition in the first case means that only if the best action (for the updated action value function) is selected, then its corresponding reward will be used to update the estimated average reward. In contrast to RVI Q-leaning, no special state needs to be selected when conducting R-learning, and hence it would potentially be more compatible with function approximation. Actually, to the best of our knowledge, \textbf{our work \cite{Our_TWC_2021} is the first study that develops DRL algorithms to maximize the long-term average reward for continuing MDPs by integrating R-learning and traditional DRL algorithms.}

Particularly, on the one hand, compared with existing DRL algorithms, we have redefined the related state value and action value functions as well as the Bellman optimality equation without introducing the discount factor (i.e., (15)-(17) in \cite{Our_TWC_2021}), which are meaningful and finite, and accordingly utilized a different equation to calculate the target value during the training (i.e., (22) in \cite{Our_TWC_2021}), which is the basis for updating the parameters of the utilized ANNs. On the other hand, to make the vanilla R-learning more compatible with the function approximation, we have made two modifications on its average reward update procedure (i.e., (\ref{Eq:R_Learning_Ave_R})). First, in our developed algorithm, the average reward is updated with the batch form without checking the condition presented in (\ref{Eq:R_Learning_Ave_R}). This makes sense since $\tilde{U}$ is introduced to estimate the average reward for the adopted policy. Nonetheless, in (\ref{Eq:R_Learning_Ave_R}) the average reward would be updated only when the greedy action is taken, which generally results in the waste of experience information and lowers down the learning efficiency, since for DRL algorithms the exploration is necessary especially at the beginning of the learning. Second, in \cite{RL_1993} the average reward is updated by resorting to the newly updated value function $R(s,a)(k+1)$ (i.e., (\ref{Eq:R_Learning})), while for our devised algorithm \cite{Our_TWC_2021}, the outputs of target ANNs, who have the much lower update frequency than other ANNs, are used to update the average reward. With this modification, it can be avoid that the update procedure keeps ``tracking'' the changing target values and hence, the stability of our proposed algorithm can be further improved.

\section{Conclusion remarks}

In this technical report, we have discussed the technical contributions of the R-learning based DRL algorithm devised in our recent study \cite{Our_TWC_2021}. Particularly, we first theoretically prove that for continuing MDPs, maximizing the discounted long-term cumulative reward is not the same as maximizing the long-term average reward, unless in the limiting regime where the discount factor $\gamma$ approaches 1. As such, in theory, the existing DRL algorithms, initially developed to optimize the discounted cumulative reward, are not suitable for maximizing the long-term average reward. Then, we have discussed why R-learning, compared with the RVI Q-leaning, is potentially more compatible with DRL algorithm design. Finally, we elaborate on what modifications we have performed on the vanilla R-learning and traditional DRL algorithms to make the proposed algorithm more compatible with the function approximation realized by utilizing ANNs. We hope this technical report will be useful for researchers who are also interested in DRL algorithm design to maximize (resp. minimize) the average reward (resp. cost) for continuing decision-making tasks.

\bibliographystyle{IEEEtran}
\bibliography{IEEEabrv,RL_UP_Ref}

\end{document}